\documentclass[sigconf]{acmart}

\usepackage{amsmath}
\usepackage{array}
\usepackage{bm} 
\usepackage{pifont}
\usepackage{mathtools}
\usepackage{graphicx} 
\usepackage{enumitem}
\usepackage{multirow}
\usepackage{subfig}
\usepackage[normalem]{ulem}
\useunder{\uline}{\ul}{}

\newcolumntype{M}[1]{>{\centering\arraybackslash}m{#1}}
\newcommand{\xmark}{\ding{55}}
\newcommand*{\boldcheckmark}{\ding{52}
}

\DeclarePairedDelimiter\autobracket{(}{)}
\newcommand{\br}[1]{\autobracket*{#1}}

\definecolor{ao(english)}{rgb}{0.0, 0.5, 0.0}
\definecolor{cadmiumred}{rgb}{0.89, 0.0, 0.13}
\newcommand{\fbseries}{\unskip\setBold\aftergroup\unsetBold\aftergroup\ignorespaces}
\newcommand{\setBoldness}[1]{\def\fake@bold{#1}}

\newlength{\Oldarrayrulewidth}
\newcommand{\Cline}[2]{%
  \noalign{\global\setlength{\Oldarrayrulewidth}{\arrayrulewidth}}%
  \noalign{\global\setlength{\arrayrulewidth}{#1}}\cline{#2}%
  \noalign{\global\setlength{\arrayrulewidth}{\Oldarrayrulewidth}}}

\AtBeginDocument{%
  \providecommand\BibTeX{{%
    \normalfont B\kern-0.5em{\scshape i\kern-0.25em b}\kern-0.8em\TeX}}}

\acmSubmissionID{1576}

\setcopyright{rightsretained}
\begin{document}
\fancyhead{}
\title{ALANET: Adaptive Latent Attention Network for\\
Joint Video Deblurring and Interpolation}

\author{Akash Gupta}
\orcid{1234-5678-9012}
\affiliation{%
  \institution{University of California, Riverside}
}
\email{agupt013@ucr.edu}

\author{Abhishek Aich}
\affiliation{%
  \institution{University of California, Riverside}
}
\email{aaich001@ucr.edu}

\author{Amit K. Roy-Chowdhury}
\affiliation{%
  \institution{University of California, Riverside}
  }
 \email{amitrc@ece.ucr.edu}


\begin{abstract}
    Existing works address the problem of generating high frame-rate sharp videos by separately learning the frame deblurring and frame interpolation modules. 
    Most of these approaches have a strong prior assumption that all the input frames are blurry whereas in a real-world setting, the quality of frames varies. Moreover, such approaches are trained to perform either of the two tasks - deblurring or interpolation - in isolation, while many practical situations call for both.
    Different from these works, we address a more realistic problem of high frame-rate sharp video synthesis with no prior assumption that input is always blurry. We introduce a novel architecture, Adaptive Latent Attention Network (ALANET), which synthesizes sharp high frame-rate videos with no prior knowledge of input frames being blurry or not, thereby performing the task of both deblurring and interpolation.   
    We hypothesize that information from the latent representation of the consecutive frames can be utilized to generate optimized representations for both frame deblurring and frame interpolation.
    Specifically, we employ combination of self-attention and cross-attention module between consecutive frames in the latent space to generate optimized representation for each frame. The optimized representation learnt using these attention modules help the model to generate and interpolate sharp frames. 
    Extensive experiments on standard datasets demonstrate that our method performs favorably against various state-of-the-art approaches, even though we tackle a much more difficult problem.
    The project page is available at \texttt{\url{https://agupt013.github.io/ALANET.html}}.
\end{abstract}


\copyrightyear{2020}
\acmYear{2020}
\acmConference[MM '20]{Proceedings of the 28th ACM International
Conference on Multimedia}{October 12--16, 2020}{Seattle, WA, USA}
\acmBooktitle{Proceedings of the 28th ACM International Conference on
Multimedia (MM '20), October 12--16, 2020, Seattle, WA, USA}
\acmDOI{10.1145/3394171.3413686}
\acmISBN{978-1-4503-7988-5/20/10}


\begin{teaserfigure}
    \centering
    \includegraphics[width=0.95\columnwidth]{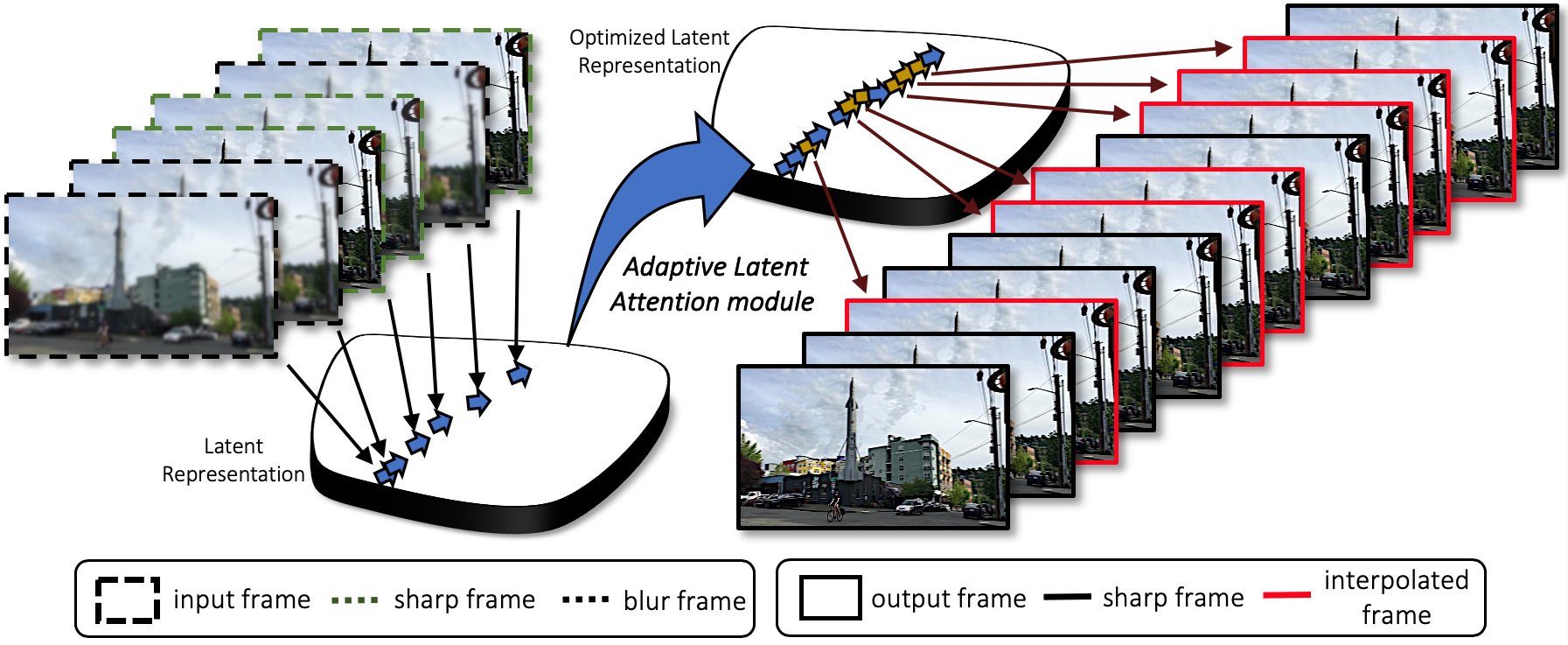}
    \caption{\textit{Conceptual Overview of} ALANET. Given a poor-quality video consisting both blurry and sharp frames, the frames are projected on a latent space. These latent representations are modulated and interpolated using the proposed Adaptive Latent Attention module to generate optimized latent representations for deblurring and interpolation. These optimized representations are then used to generate a high frame-rate sharp video.}
    \label{fig:concept}
\end{teaserfigure}

\maketitle

\section{Introduction}
Motion blur and low frame-rate are often commonplace in videos captured by mobile devices, whether hand-held or on a moving platform. The reasons vary, including low shutter frequency, long exposure times, and the movement of the device itself~\cite{telleen2007synthetic, jin2018learning}. 
These factors limit the quality of videos captured.
As vast majority of video media is captured using mobile cameras these days, it calls for improved quality of the videos captured by these devices. Enhancing video quality requires restoring the degradation caused by motion blur along with increase in the frame-rate at which video is captured for temporal smoothness.

Most existing approaches have addressed the problem of high frame-rate sharp video generation by frame deblurring and frame interpolation, separately.
In~\cite{jin2018learning}, separate models are used to deblur input frames and to interpolate between frames. The phenomenon of motion blur and frame-rate at which video is captured are related. Thus, a joint formulation is needed when addressing the task of high frame-rate sharp video generation from a low frame-rate blurry video.
Recently, \cite{shen2020blurry} studied the problem of joint video deblurring and interpolation. Here, authors proposed to use pyramid deep models to deblur and interpolate along with a pyramid of convolutional Long-Short Term Memory (LSTM) to capture temporal smoothness. 
However, these methods assume that all the input frames are blurry, which is often unrealistic because the quality of a video usually varies non-uniformly over time.

In this paper, we introduce a novel architecture \textbf{A}daptive \textbf{L}atent \textbf{A}ttention \textbf{NET}work (\textbf{ALANET}) which aims to jointly deblur and interpolate frames from a poor quality video input without an assumption that all input frames are blurry. Specifically, we construct a Adaptive Latent Attention module that leverages the latent space with attention mechanisms to generate high frame-rate sharp video. 
\textbf{ALANET} has a U-Net variant~\cite{ronneberger2015u} as it's backbone, combined with the proposed attention module. Similar to U-Net, we utilize contracting path (encoder) of the network for latent space representation and expanding path (generator) for video generation. However unlike U-Net, we do not pass the bottleneck features extracted from the encoder directly to the generator. We introduce our proposed adaptive attention module to modulate and interpolate the latent features for deblurring and interpolating frames from the input video. Figure~\ref{fig:concept} illustrates the concept of proposed adaptive attention module. Given a set of input blurry and sharp frames, their projection in latent space can be modulated and interpolated using Adaptive Latent Attention module, to generate optimized representations for sharp frames. These modulated and interpolated latent representations are then used by the generator to synthesize the high frame-rate sharp video.

\subsection{Approach Overview} 
\begin{figure*}[t]
    \centering
    \includegraphics[width=0.925\textwidth]{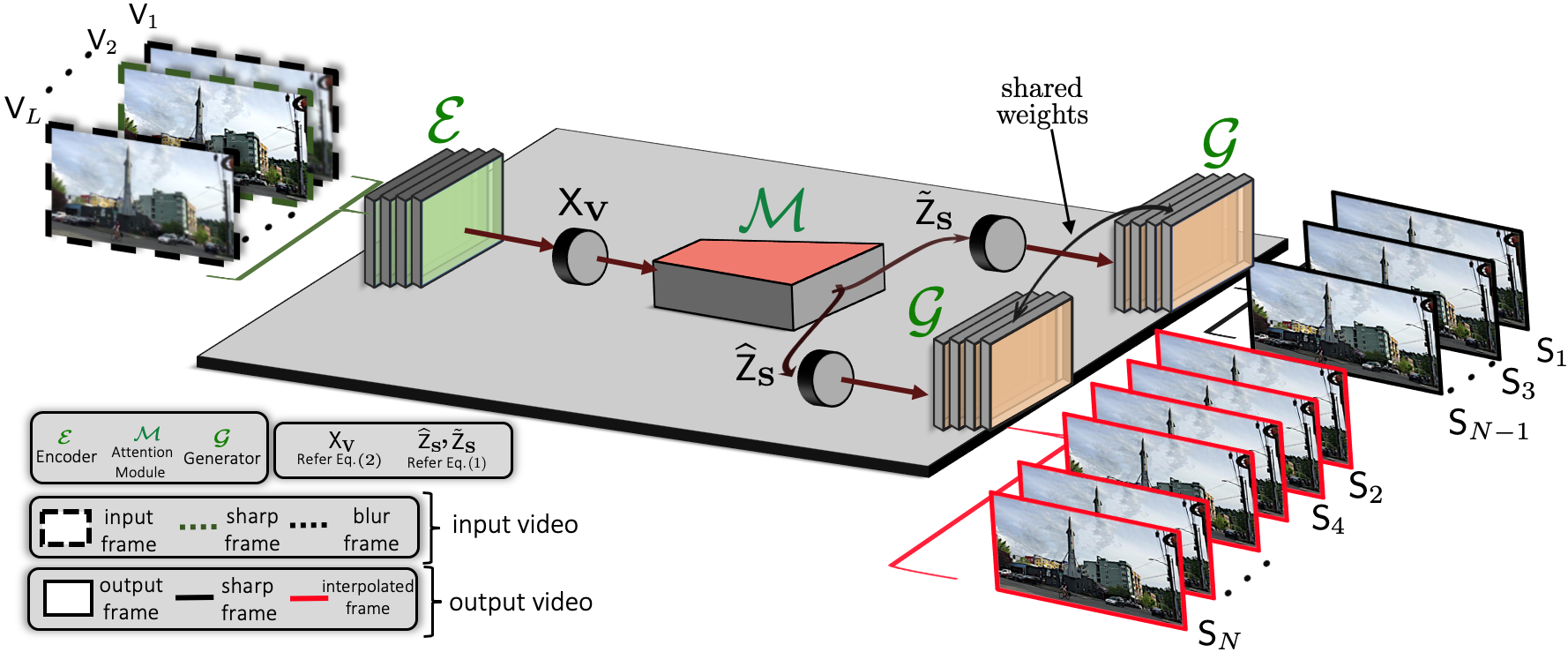}
    \caption{\textit{Architectural Overview of} ALANET. Given a low frame-rate poor quality video  $ \textbf{V} = [~\mathsf{V}_1, ~\mathsf{V}_2, \cdots,~\mathsf{V}_L ]$, we extract latent representations $\mathsf{X}_{\textbf{V}} = [ ~\textbf{x}_1, ~\textbf{x}_2, \cdots, ~\textbf{x}_L ~ ]$ using encoder network $\mathcal{E}$. Adaptive Latent Attention module $\mathcal{M}$ utilizes combination of self-attention and cross-attention on  $\mathsf{X}_{\textbf{V}}$ to generate optimized representations for deblurring ($\widetilde{\mathsf{Z}}_\textbf{S}$) and interpolation ($\widehat{\mathsf{Z}}_\textbf{S}$). These optimized representations are used by the generative network $\mathcal{G}$ to synthesize deblurred frames ($\mathsf{S_1}, \mathsf{S_3}, \cdots, \mathsf{S}_{N-1}$) from $\widetilde{\mathsf{Z}}_\textbf{S}$ and interpolated frames ($\mathsf{S_2}, \mathsf{S_4}, \cdots, \mathsf{S}_{N}$) from  $\widehat{\mathsf{Z}}_\textbf{S}$, thereby generating a high frame-rate video $\textbf{S} = [ \mathsf{S}_1, ~\mathsf{S}_2, \cdots, ~\mathsf{S}_N ]$.}
    \label{fig:overview}
\end{figure*}

An overview of our approach is illustrated in Figure~\ref{fig:overview}.
%
Given a low frame-rate poor quality input, our objective is to generate a high frame-rate sharp video.
%
Our proposed architecture, \textbf{ALANET}, consists of three modules: the frame encoding network $\mathcal{E}$, the Adaptive Latent Attention network $\mathcal{M}$, and the high frame-rate sharp video generator $\mathcal{G}$.  
We modulate and interpolate the frame features by applying \textbf{self-attention} and \textbf{cross-attention} on channels of the latent features of consecutive frames using our proposed adaptive attention module. Self-attention on the feature space helps the model to focus on important features of the same frame whereas cross-attention helps the model to retrieve information from neighbouring frames that can be useful for either deblurring or interpolation tasks. 
In turn, the Adaptive Latent Attention module will give less importance to the neighbouring frame feature if the input is a sharp frame, and utilize this information from the neighbours if input frame is blurry. 
Hence, our proposed approach is able to deblur and generate high quality interpolated frames using self-attention and cross-attention on frame representations. To the best of our knowledge, \emph{our approach is the first work to exploit the ability of learning optimized latent representation for generation of high frame-rate sharp video using self-attention and cross-attention}. 

\subsection{Contributions}
The key contributions of our proposed framework are summarized as follows.

\begin{itemize}[leftmargin=*]
    \setlength \itemsep{0.4em}
     \item We introduce a novel framework \textbf{ALANET}, Adaptive Latent Attention Network, designed to jointly deblur and interpolate for high frame-rate visually sharp video generation. 
     \item This is the first work to generate high frame-rate sharp video from low frame-rate poor quality video by applying attention in the latent space without any assumption on the uniformity of blurriness in different frames of the video.    
     \item Our framework demonstrates consistently effective results on two datasets, the benchmark Adobe240 and crawled YouTube240 with better or at par performance with state-of-the-art in both deblurring and interpolation tasks. 
\end{itemize}

\section{Related Work}
Our work relates to research in video deblurring, video interpolation, attention model, and joint video deblurring and interpolation. In this section, we discuss some representative methods closely related to our work (see Table \ref{tab:compare_methods}).
\bigskip

\noindent \textbf{Video Deblurring.} Inversion of motion blur is an ill-posed problem~\cite{raskar2006coded, nimisha2017blur}. Recent works have used deep learning based methods to solve this restoration problem either using a single frame~\cite{su2017deep, tao2018scale} or multiple frames~\cite{jin2018learning, kim2017dynamic, hyun2015generalized, su2017deep, nah2017deep}. \cite{cho2012video} attempts to deblur a video by exploring similarity between the frames of the video and exploiting sharp patches of neighbouring frames. DeBlurNet \cite{su2017deep} proposes to use consecutive frames stacked as input to generate a single clean central frame. ESVR~\cite{wang2019edvr} tries to align the features of multiple frames using a temporal and spatial fusion module for feature fusion from different layer to deblur a video. \cite{kim2016dynamic} proposes an integrated model to jointly predict the defocus blur, optical flow and latent frames. \cite{hyun2017online} proposed a spatio-temporal recurrent neural network that enforces temporal consistency between neighbouring frames. \cite{zhou2019spatio} proposes a spatio-temporal recurrent architecture with dynamic temporal blending mechanism. In contrast, we do not estimate any extra information like optical flow (which can be noisy and computationally heavy) in our approach  and rely on proposed attention model to generate high frame-rate sharp videos.\medskip

\noindent \textbf{Video Interpolation.} Many of the existing approaches \cite{mahajan2009moving, zitnick2004high, bao2019depth, bao2019memc, jiang2018super, liu2017video} for frame interpolation use optical flow estimation between input frames. Consequently, the quality of estimated optical flow governs the quality of frame interpolation. Recent learning based methods have demonstrated  effectiveness in frame interpolation tasks. A direct application of convolutional neural networks (CNNs) for intermediate frame synthesis is presented in~\cite{long2016learning}. Some methods \cite{niklaus2017video, niklaus2017video2} apply CNNs to estimate space-varying and separable convolutional kernels for synthesis using neighbourhood pixels. \cite{aich2020non} proposes to generate videos by learning optimized representation by a non-adversarial approach and then interpolating between the optimized latent representation of two frames to synthesize central frame. However, they average the latent representations of two frames for frame interpolation which often generates a blurry image. Unlike these methods, our approach utilizes adaptive attention in the latent space for interpolation.\medskip

\noindent \textbf{Attention Model.} Attention mechanism has garnered a lot of interest due to their learnable guidance ability. With pioneering work in language translation~\cite{vaswani2017attention}, variations of attention mechanism have shown promising results in object recognition~\cite{ba2014multiple}, image generation~\cite{zhang2018self} and image super-resolution~\cite{zhang2018image}. Residual channel attention mechanism for super-resolution is introduced in \cite{zhang2018image}. Authors in~\cite{wu2020david} used different length sequences to deblur the center frame and attention is applied on different outputs to generate a single central frame. Recently, variations of attention models are proposed for video deblurring~\cite{wu2020david} and video interpolation~\cite{choi2020channel}. In \cite{choi2020channel}, attention is applied channel-wise on concatenated down-shuffled frames for video interpolation. In contrast to our work, where we apply attention in latent space, the existing methods employ attention for video deblurring and interpolation tasks in pixel space.

\noindent \textbf{Joint Video Deblurring and Interpolation.} Joint video deblurring and interpolation still remains a challenging problem. \cite{jin2018learning} proposed DeBlurNet, to deblur, and InterpNet, for interpolating input frames in a jointly optimized cascade scheme to generate sharp slow motion videos using blurry input. Blurry Video Frame Interpolation proposed in \cite{shen2020blurry} uses pyramid structure to deblur and interpolate along with a pyramid convolutional LSTM to capture temporal information. However, both these methods strongly assume that all the input frames are blurry. We relax this assumption to address a more difficult problem where we do not know which input frames are blurry and where to interpolate.
Hence, the proposed \textbf{ALANET} framework is \emph{self-sufficient to make decisions on which frames to deblur using information from neighbouring frames.}

\begin{table}[H]
\caption{\textit{Categorization of prior works in video deblurring and interpolation}. Different from the state-of-the-art approaches, ALANET demonstrates adaptive attention in latent space to perform joint deblurring and interpolation.}

\begin{center}
\renewcommand{\arraystretch}{1.25}
\begin{tabular}{M{1.3cm}|M{1.3cm}|M{0.9cm}|M{1.8cm}|M{1.3cm}}
\toprule[1.2pt]
\multicolumn{1}{c|}{\multirow{3}{*}{Methods}} & \multicolumn{4}{c}{Settings}\\
\cline{2-5}
& \small{Interpolate?} & \small{Deblur?} & \small{Joint Deblur \& Interpolate?} & \small{Latent Attention?}\\ %
\midrule
DAIN~\cite{bao2019depth}         & \textcolor{ao(english)}{\boldcheckmark}   & \textcolor{cadmiumred}{\xmark}            & \textcolor{cadmiumred}{\xmark}            & \textcolor{cadmiumred}{\xmark} \\
\hline
Jin~\cite{jin2018learning}       & \textcolor{ao(english)}{\boldcheckmark}   & \textcolor{ao(english)}{\boldcheckmark}   & \textcolor{cadmiumred}{\xmark}            &  \textcolor{cadmiumred}{\xmark} \\
\hline
BIN~\cite{shen2020blurry}        & \textcolor{ao(english)}{\boldcheckmark}   &  \textcolor{ao(english)}{\boldcheckmark}  &  \textcolor{ao(english)}{\boldcheckmark}  & \textcolor{cadmiumred}{\xmark}  \\
\hline
\textbf{ALANET} (Ours)           & \textcolor{ao(english)}{\boldcheckmark}   & \textcolor{ao(english)}{\boldcheckmark}   & \textcolor{ao(english)}{\boldcheckmark}   & \textcolor{ao(english)}{\boldcheckmark}\\
\bottomrule[1.2pt]

\end{tabular}
\end{center}
\label{tab:compare_methods}

\end{table}
%
%
%
%
\section{Problem Formulation}
\label{sec:formulation}

Given a low frame-rate poor quality video 
    $ \textbf{V} = [ ~\mathsf{V}_1, ~\mathsf{V}_2, \cdots,~\mathsf{V}_L ]$,
with $L$ frames, we aim to generate a high frame-rate sharp video
    $\textbf{S} = [ \mathsf{S}_1, ~\mathsf{S}_2, \cdots, ~\mathsf{S}_N ]$ 
with $N$ frames, where $N > L$. %
Our objective is to deblur and increase the frame-rate of the given input video $\textbf{V}$. Corresponding to each input frame $~\mathsf{V}_i~\forall~i = ~1,~2, \cdots,~L$, ~~let there be a feature representation $\textbf{x}_i$ in latent space  $\mathsf{X}\in\mathbb{R}^{H_1 \times W_1 \times C_1 \times L}$ such that $
    \mathsf{X}_{\textbf{V}} = [ ~\textbf{x}_1, ~\textbf{x}_2, \cdots, ~\textbf{x}_L ~ ]
$
where $H_1 \times W_1 \times C_1$ is the dimension of the latent representation.

We propose to generate a high frame-rate video by adaptive attention modeling (see Section~\ref{ssec:attn})
of the feature representations of input video frames in the latent space. Our hypothesis is that in latent space, information from neighbouring frames can help learn optimized representations for deblurring and interpolation. Thus, the proposed Adaptive Latent Attentive model transforms input blurry frame representation ($\mathsf{X}_{\textbf{V}}$) to the optimized representations ($\mathsf{Z}_{\textbf{S}}\in\mathbb{R}^{H_1\times W_1 \times C_1 \times N}$) for deblurring and interpolation in the latent space given by
\begin{align}
    \mathsf{Z}_{\textbf{S}} = [ \textbf{z}_1,~\widehat{\textbf{z}}_2,~\textbf{z}_3,~\widehat{\textbf{z}}_4,\cdots,~\textbf{z}_N ] =  \widetilde{\mathsf{Z}}_{\textbf{S}} \textstyle \bigcup \widehat{\mathsf{Z}}_{\textbf{S}}
\end{align}
where $\textbf{z}_{2i}$ is the representation for a deblurred frame $~\mathsf{S}_{2i}$, and $~\widehat{\textbf{z}}_{2i+1}$ is the representation for an interpolated frame between $~\mathsf{S}_{2i}$ and $~\mathsf{S}_{2i+2}$, i.e., $~\mathsf{S}_{2i+1}$. We denote all latent representations for deblurred frames by $\widetilde{\mathsf{Z}}_{\textbf{S}}$ and for interpolated frames by $\widehat{\mathsf{Z}}_\textbf{S}$. These optimized representations $\mathsf{Z}_{\textbf{S}} = \widetilde{\mathsf{Z}}_{\textbf{S}} \bigcup \widehat{\mathsf{Z}}_{\textbf{S}}$ are used to deblur and interpolate sharp frames to generate a high frame-rate video. 

\section{ALANET: Adaptive Latent Attention Network}
\label{sec:Alanet}
In this section, we describe the proposed framework, \textbf{ALANET}, in detail. Our framework consists of three components: the encoder $\mathcal{E}$, the Adaptive Latent Attention module $\mathcal{M}$ and, the generator $\mathcal{G}$. We use the encoder module to extract latent representation for each input frame. The Adaptive Latent Attention module generates optimized representations for frames to reduce blur and to interpolate frames, simultaneously. Finally, the optimized representations are used by the generator to synthesize a high frame-rate sharp video. Our overall framework is shown in Figure~\ref{fig:overview}.

\subsection{Latent Representation of Frames}
The encoder $\mathcal{E}$ is a trainable convolutional neural network which projects the input video into a latent representation for each frame. 
\begin{alignat}{3}
    \mathcal{E}\br{\textbf{V}} 
    &= \mathcal{E}\Big{(} [ ~\mathsf{V}_1, ~\mathsf{V}_2, \cdots, ~\mathsf{V}_L ]\Big{)} \\ \nonumber
    &= [~\textbf{x}_1, ~\textbf{x}_2, \cdots, ~\textbf{x}_L  ]
     = \mathsf{X}_{\textbf{V}}
\end{alignat}
Here, $\textbf{x}_i \in\mathbb{R}^{H_1 \times W_1 \times C_1}$ is the latent representation corresponding to $\textsf{V}_i$. The representations generated by the encoder $\mathcal{E}$ are used by the Adaptive Latent Attention module $\mathcal{M}$ to generate optimized representations for deblurring and interpolation.

\subsection{Adaptive Latent Attention}
\label{ssec:attn}
The latent representation of a frame generated by the encoder may not be optimized as all the channels of the input representation are not equally important for generation task. Also, since frames of a video are temporally correlated, their latent representation can be leveraged to extract information from neighbouring frames to generate an optimized representation for deblurring and interpolation. 

To extract important information from the latent representation of the given frame and utilize the information from the neighbouring frames, we propose an Adaptive Latent Attention 
module $\mathcal{M}$. The proposed module $\mathcal{M}$ applies attention on the input latent representations to generate the optimized representations for deblurring and interpolation. This module takes two latent representations ($\textbf{x}_i$, $\textbf{x}_j \in \mathbb{R}^{H_1 \times W_1 \times C_1}$) as input, where $H_1 \times W_1$ is dimension of each feature in $C_1$ channels of the latent representation. 
A combination of \textbf{self-attention} $\mathcal{M}_S$ and \textbf{cross-attention} $\mathcal{M}_C$ is then used to generate latent representations to jointly deblur and interpolate between two consecutive frames in an adaptive manner. 

\begin{figure}[t]
    \centering
    \subfloat[Attention Mechanism]{\includegraphics[width=0.99\columnwidth]{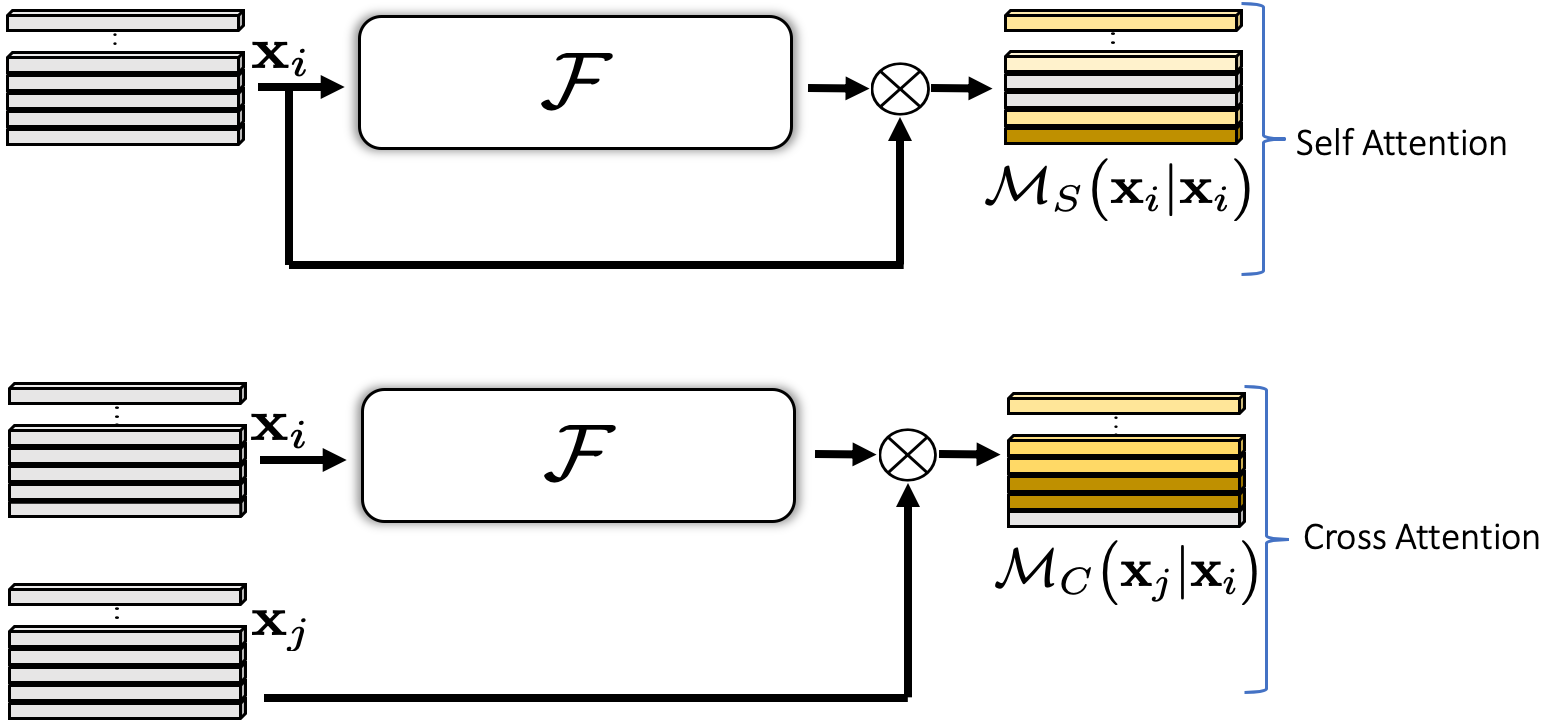}}
    \hfill
    \subfloat[Channel Attention Computation Network]{\includegraphics[width=0.99\columnwidth]{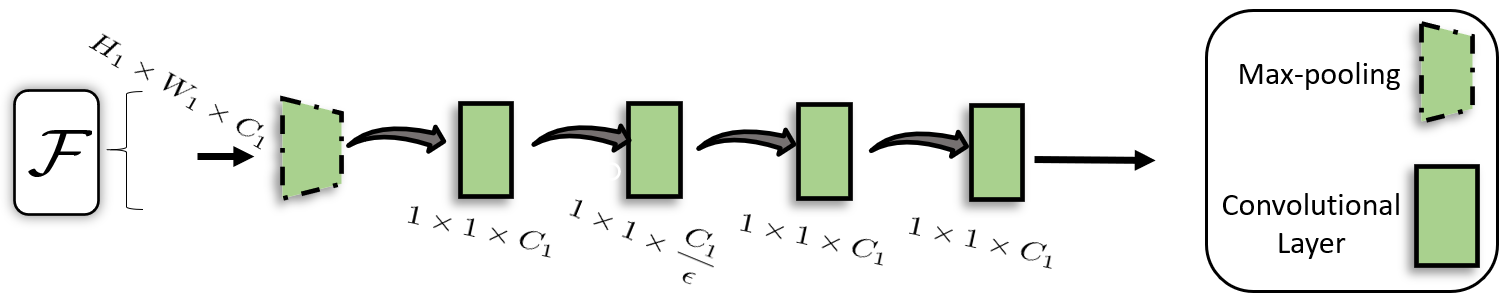}}
    \caption{\textit{Proposed Attention Module.} (a) Self-Attention (\textit{top}) on latent representation $\textbf{x}_i$ and Cross-Attention (\textit{bottom}) for representation $\textbf{x}_j$ conditioned on $\textbf{x}_i$. Symbol $\otimes$ denotes element-wise multiplication of each attention weight with respective channel of the representation. (b) The channel weight computation function $\mathcal{F}$. It generates channel descriptor by channel-wise global average pooling to learn attention weights for each channel.}
    \label{fig:attention}
\end{figure}

    
  

The basic building block of the attention mechanism is the channel attention function $\mathcal{F}$. It computes attention weights of each channel in the latent representation. As in~\cite{zhang2018image}, the channel-wise global spatial information is extracted using global average pooling to condense input features to a channel descriptor. Then, a gating mechanism is applied to learn non-linear interactions 
and correlation between multi-channel features such that $\mathcal{F}:\mathbb{R}^{H_1\times W_1 \times C_1} \to \mathbb{R}^{1\times 1 \times C_1}$, where $H_1 \times W_1 \times C_1$ is the dimension of the latent representation. Figure~\ref{fig:attention} shows the self-attention $\mathcal{M}_S$ and cross-attention $\mathcal{M}_C$ modules along with the basic building block $\mathcal{F}$ for computation of the channel attention.\bigskip

\noindent \textbf{Self-Attention} ($\mathcal{M}_S$) correlates different channels of the latent representation of a frame in order to generate an informative representation. This is achieved by computing attention weights for each of the channels of the input representation followed by element-wise multiplication of the channels with their attention weights. This self-attention on $\textbf{x}_i$ can then be expressed as in~\eqref{eqn:self_attn}.\bigskip

\noindent \textbf{Cross-Attention} ($\mathcal{M}_C$) provides attention weights for each channel of the latent representation $\textbf{x}_j$ conditioned on another latent representation $\textbf{x}_i$. Cross-attention leverages information from other frames to generate a conditional representation. The conditional representation provides insight on what information is useful from other frames. This cross-attention on $\textbf{x}_j$ given the input $\textbf{x}_i$ can then be computed as in~\eqref{eqn:cross_attn}.
\begin{align}
    \mathcal{M}_S\Big{(}\textbf{x}_i\vert \textbf{x}_i\Big{)} = \textbf{x}_i \otimes \mathcal{F}(\textbf{x}_i) 
    \label{eqn:self_attn} 
\end{align}
\begin{align}
    \mathcal{M}_C\Big{(}\textbf{x}_j \vert \textbf{x}_i\Big{)} = \textbf{x}_j \otimes \mathcal{F}(\textbf{x}_i)
    \label{eqn:cross_attn}
\end{align}
\noindent Note that, $\otimes$ in \eqref{eqn:self_attn} and \eqref{eqn:cross_attn} represents element-wise multiplication, $\textbf{x}_i, \textbf{x}_j~\in ~\mathbb{R}^{H_1 \times W_1 \times C_1}$ are the encoded feature representations of frames and $\mathcal{M}_\textbf{S}(\textbf{x}_i \vert \textbf{x}_i),~\mathcal{M}_\textbf{C}(\textbf{x}_j \vert \textbf{x}_i) ~\in ~\mathbb{R}^{H_1 \times W_1 \times C_1}$.\bigskip

\noindent \textbf{Deblurred and Interpolated Representations.}
A combination of self-attention and cross-attention modules is employed to obtain optimized latent representations for deblurring and interpolation. Given a window $\mathbf{W}$, the optimized latent representations 
$ \mathsf{Z}_{\textbf{V}} = [~\textbf{z}_1,~\widehat{\textbf{z}}_2,~\textbf{z}_3,~\widehat{\textbf{z}}_4, \cdots,~\textbf{z}_N  ]$ for a high frame-rate video $\textbf{S}$ is computed as follows: 
\begin{alignat}{3}
    \textbf{z}_{2i} 
    &= 
    \mathcal{M}_S\Big{(}\textbf{x}_i\vert \textbf{x}_i\Big{)} + \sum_{\mathclap{j\in\mathbb{Q}}} 
    \mathcal{M}_C\Big{(}\textbf{x}_j\vert \textbf{x}_i\Big{)}\label{eqn:deblur} 
     \\ \nonumber
   \widehat{\textbf{z}}_{2i + 1} 
   &= 
   \mathcal{M}_S\Big{(}\textbf{x}_i\vert \textbf{x}_i\Big{)} +
   \mathcal{M}_C\Big{(}\textbf{x}_i\vert\textbf{x}_{i+1}\Big{)} \\
    &\quad+ 
    \mathcal{M}_S\Big{(}\textbf{x}_{i+1}\vert \textbf{x}_{i+1}\Big{)} + 
    \mathcal{M}_C\Big{(}\textbf{x}_{i+1}\vert\textbf{x}_i\Big{)}
    \label{eqn:interp}
\end{alignat}
\noindent where $\mathbb{Q}$ denotes integer values in $[~i - 0.5\mathbf{W},~i~) ~\bigcup ~(~i,~i + 0.5\mathbf{W}~]$,
$\textbf{z}_{2i}$ is the optimized representation for deblurred frame $\mathsf{S}_{2i}$ and $\widehat{\textbf{z}}_{2i + 1}$ is the optimized representation for the interpolated frame between $\mathsf{S}_{2i}$ and $\mathsf{S}_{2i + 2}$. 

As defined by \eqref{eqn:deblur}, an optimized representation $\textbf{z}_{2i}$ for sharp output $\mathsf{S}_{2i}$ is computed using self-attention on $i^{th}$ input representation $\textbf{x}_i$ and cross-attention of all the remaining input latent representation $\textbf{x}_j$ in a neighbourhood of $\mathbf{W}$ frames. Cross-attention is computed in a temporal window of $\mathbf{W}$ frames as the significant information for deblurring and interpolation is available in neighbouring frames compared to temporally distant frames.
Similarly, a latent representation $\widehat{\textbf{z}}_{2i+1}$ for interpolated frame $\mathsf{S}_{2i + 1}$ between $\mathsf{S}_{2i}$ and $\mathsf{S}_{2i+2}$ is given by ~\eqref{eqn:interp}, where we consider self-attention on each latent representations $\textbf{x}_{i}$ and $\textbf{x}_{i+1}$, and cross-attention for each representation conditioned on the other.

\subsection{High Frame-Rate Video Generation}
To generate a high frame-rate video from blurry inputs, we employ a generative neural network $\mathcal{G}$ that transforms the optimized representations to a sequence of frames. The optimized representations generated by the adaptive attention module $\mathcal{M}$ are used by generator $\mathcal{G}$ to synthesize deblurred frames as well as interpolate between frames represented by $\textbf{S}= \mathcal{G}\br{ [ ~\textbf{z}_1, ~\widehat{\textbf{z}}_2, ~\textbf{z}_3, ~\widehat{\textbf{z}}_4,\cdots, ~\textbf{z}_N   ]}$  where $\textbf{z}_{2i}$ and $\widehat{\textbf{z}}_{2i+1}$ are optimized representation used to deblur and interpolate frames $\mathsf{S}_{2i}$ and $\mathsf{S}_{2i+1}$, respectively.

\subsection{Network Architecture}
In this section, we describe the network architecture used for different modules in the proposed \textbf{ALANET} framework.\smallskip

\noindent \textbf{Encoder-Generator Network}. A variation of U-Net \cite{jiang2018super} is employed to design the backbone network for the proposed framework. The contracting path is used as the encoder network $\mathcal{E}$ and the expansive path is used as the generator network $\mathcal{G}$. The encoder-decoder network also retains the skip-connections as in the original U-Net architecture~\cite{ronneberger2015u}. However unlike the U-Net architecture, our proposed Adaptive Latent Attention module $\mathcal{M}$ is introduced after the bottleneck to optimize the latent representations before they are fed to the generator $\mathcal{G}$.\smallskip

\noindent \textbf{Adaptive Latent Attention Network.} In order to make the generator model, $\mathcal{G}$, focus more on informative features, we exploit the inter-dependencies within frame feature (self-attention) and across frame features (cross-attention). The basic building block of self-attention and cross-attention is the attention weight computation module, $\mathcal{F}$. We adopt the channel attention module as in \cite{zhang2018image} for $\mathcal{F}$. This channel attention module first extracts the channel-wise  global spatial information into a channel descriptor using global average pooling. Then, a gating mechanism is applied to learn non-linear interactions and non-mutually-exclusive relationship between multi-channel features~ \cite{zhang2018image}. 
Unlike self-attention for super-resolution in~\cite{zhang2018image}, we also employ cross-attention between consecutive features to learn interactions between these features for deblurring and interpolation.

\begin{figure*}[htp]
\subfloat[\textit{Representative result from} Adobe240 \textit{dataset}. Observe zoomed-in patch of the car.
The motion of car introduces motion blur. ALANET is able to significantly reduce the motion blur in all the frames and also generate superior quality interpolated frames.]
{\label{subfig:compare_a}
  \includegraphics[clip,width=0.95\linewidth]{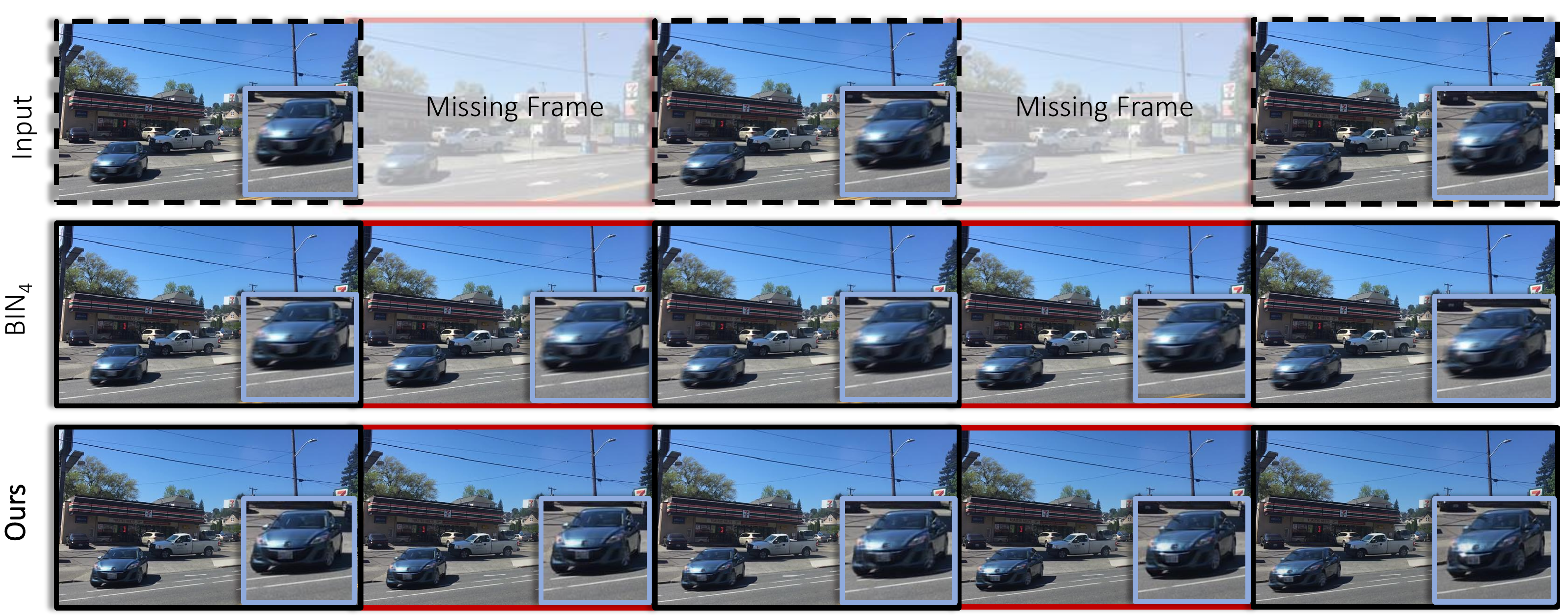}%
}
\vspace{-0.7em}
\subfloat[\textit{Representative result from} Adobe240 \textit{dataset}. The last frame in blurry input (top row) is of poor quality. ALANET is able to deblur and interpolate clear frame (last two frame in the bottom row) as compared to the state-of-the-art (last two frame in the middle row).]
{ \label{subfig:compare_b}

  \includegraphics[clip,width=0.95\linewidth]{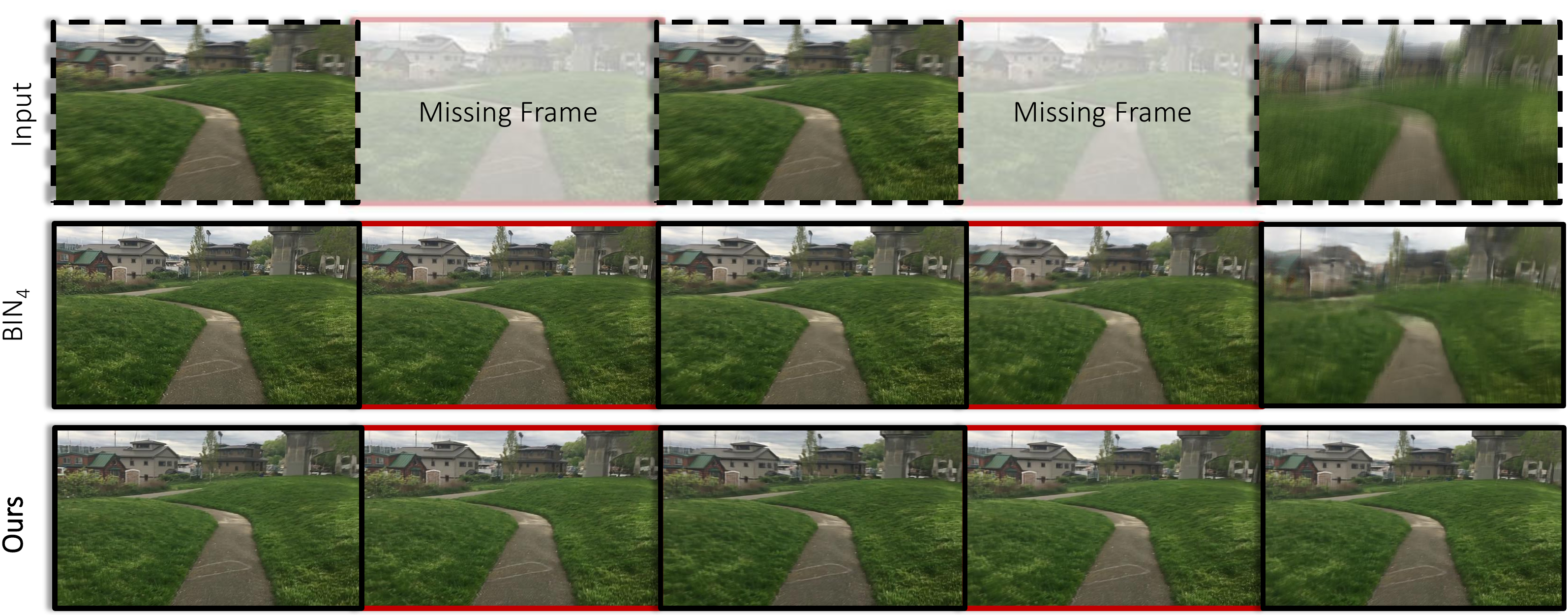}%
}
\vspace{-0.7em}
\caption{\textit{Qualitative result comparison with the state-of-the-art.} Top row consists of the input blurry frames and the missing frames faded. We show two high frame-rate videos generated by our proposed method (bottom row) and compare it with the state-of-the-art BIN$_4$ (middle row). ALANET is able to generate superior quality high frame-rate video. } 
\label{fig:compare}
\end{figure*}

\section{Experiments}
In this section, we first introduce the benchmark datasets, and evaluation metrics. Next, the model used for generation of blurry training data is described. Finally, extensive experiments are shown to demonstrate the effectiveness of our proposed approach in generating high frame-rate sharp videos.

\subsection{Datasets and Metrics}
We evaluate the performance of our approach using publicly available Adobe240~\cite{su2017deep} dataset which has been used in many prior works and a dataset crawled from YouTube as in~\cite{shen2020blurry}.\medskip

\noindent \textbf{Adobe240 Dataset.} This dataset contains 118 videos captured at 240 frames per second (fps) with the resolution of $1280 \times 720$ . We choose 110 videos for training and remaining 8 for evaluation following the split provided in~\cite{jiang2018super} for fair comparison.\medskip

\noindent \textbf{YouTube240 Dataset.} We download 60 random video videos captured at 240fps from the YouTube website to construct an evaluation dataset similar to that used in~\cite{shen2020blurry}. The resolution of the downloaded video is $1280 \times 720$. For this dataset, we train the model in Adobe240 but test on YouTube240 without any fine-tuning.\medskip

\noindent \textbf{Dataset Preparation.} For Adobe240~\cite{su2017deep} and crawled YouTube240 dataset, low frame-rate poor quality videos of 30fps are generated using process described in section~\ref{ssec:implement}. All the frames are resized to $640\times352$ for training and evaluation purposes.

\subsection{Implementation Details}
\label{ssec:implement}
Our framework is implemented in PyTorch \cite{paszke2017automatic}. All the experiments are trained for 200 epochs with a batch size of 2. We use ADAM~\cite{kingma2014adam} optimizer with initial learning rate of 0.0001 and weight decay $5 \times 10^{-4}$. The learning rate is reduced by a factor of 10 after 100 and 150 epochs. 
The proposed framework takes a 30fps blurry video as an input and generates a 60fps sharp video.\medskip

\begin{table*}[t]
    \caption{\textit{Quantitative results comparison on} Adobe240 \textit{and} YouTube240. We obtained better average PSNR and SSIM index on Adobe240 dataset. Our proposed approach performs at-par on YouTube240 dataset when evaluated using the model trained on Adobe240. Best scores have been highlighted in bold. $^\dagger$ indicates results reported from~\cite{shen2020blurry}.}
    \vspace{-1em}
    \begingroup
    \renewcommand{\arraystretch}{1.435}
    \begin{tabular}{l c c c c | c c c c | c c c c }
        \toprule[0.2em]
        \multirow{3}{*}{\textbf{Method}} & \multicolumn{4}{c }{\textbf{Deblurring}}                                & \multicolumn{4}{c }{\textbf{Interpolation}}                             & \multicolumn{4}{c}{\textbf{Joint Deblurring and Interpolation}}                         \\\Cline{0.5pt}{2-13}
                        \rule{0pt}{1.5em}
                        & \multicolumn{2}{c}{Adobe240} & \multicolumn{2}{c|}{YouTube240} & \multicolumn{2}{c}{Adobe240} & \multicolumn{2}{c|}{YouTube240} & \multicolumn{2}{c}{Adobe240} & \multicolumn{2}{c}{YouTube240} \\\Cline{1pt}{2-13}
                        \rule{0pt}{1.5em}
                                    & PSNR         & SSIM          & PSNR          & SSIM           & PSNR          & SSIM         & PSNR           & SSIM          & PSNR          & SSIM         & PSNR           & SSIM          \\
                       \bottomrule[0.1em] 
                      
        Blurry Inputs$^{\dagger}$                      & 28.68        & 0.8584        & 31.96         & 0.9119         & -             & -            & -              & -             & -             & -            & -              & - \\
        Super SloMo$^{\dagger}$~\cite{jiang2018super}  & -            & -             & -             & -              & 27.52         & 0.8593       & 30.84          & 0.9107        & -             & -            & -              & - \\
        MEMC-Net$^{\dagger}$~\cite{bao2019memc}                   & -            & -             & -             & -              & 30.83         & 0.9128       & 34.91          & 0.9596        & -             & -            & -              & - \\
        DAIN$^{\dagger}$~\cite{bao2019depth}                       & -            & -             & -             & -              & 31.03         & 0.9172       & 35.09          & 0.9615        & -             & -            & -              & - \\
        \hline
        Jin$^{\dagger}$~\cite{jin2018learning}         & 29.40        & 0.8734        & 32.06         & 0.9119         & 29.24         & 0.8754       & 32.24          & 0.9140        & 29.32         & 0.8744        & 32.15          & 0.9130 \\
        \hline
        BIN$_{4}$$^{\dagger}$~\cite{shen2020blurry}    &{\ul 32.67}    & {\ul 0.9236}        & 35.10         & 0.9417         & {\ul 32.51}         & {\ul 0.9280}       & 35.10          & 0.9468        & {\ul 32.59}         & {\ul 0.9258}        & 35.10          & 0.9443 \\
        \textbf{ALANET} (Ours)               &\textbf{33.71} &\textbf{0.9429}      &35.94          & 0.9496         & \textbf{32.98}      & \textbf{0.9362}    & 35.85          & 0.9513        & \textbf{33.34}      & \textbf{0.9355}     & 35.89          & 0.9504 \\
        \bottomrule[0.2em]
    \label{tab:main}
    \end{tabular}
    \endgroup
    \vspace{-1.5em}
\end{table*}

\noindent \textbf{Blurry Video Formation.}
Camera shutter frequency affects degradation due to motion blur in each frame of a captured video. A low shutter frequency may not be able to capture temporal smoothness and hence generate blurry frames. To simulate the motion blur, we approximate the blurry frame as a discrete averaging of sharp frames within an overlapping window as defined in~\cite{jin2018learning, jiang2018super, su2017deep}. Let $2\tau + 1$ be the number of sharp frames between two blurry frames and $\beta$ be the rate at which frames are captured. Then, a blurry frame $\mathsf{V}_i$ is approximated as: 
\begin{align}
 \label{eqn:blur}
    \mathsf{V}_i = \dfrac{1}{2\tau + 1}\sum_{k = i\beta - \tau  }^{i\beta + \tau} \mathsf{S}_{k}
\end{align}
 where, $~\mathsf{S}_{k}$'s are the sharp frames in the given video. Since we do not assume that all the input frames are blurry, we average 11 consecutive frames randomly using ~\eqref{eqn:blur} on a sharp video to generate a poor quality video with low frame-rate.

\noindent \textbf{Training and Testing Protocol.}
During training, random blurry frames are generated on-the-fly by averaging 11 frames as defined in ~\eqref{eqn:blur}. The $5^{th}$ and $9^{th}$ sharp frames are considered as the ground-truth for deblurring and interpolation, respectively. The framework is jointly optimized for deblurring and interpolation using Adaptive Latent Attention Network. During testing, a low frame-rate (30fps) poor quality video is used as an input to the trained model and a high frame-rate (60fps) sharp video is generated.\bigskip

\noindent \textbf{Objective Function.} 
Our objective function consists of a $\ell_1$ pixel reconstruction loss\footnote{For pixel reconstruction loss, we choose $\ell_1$-loss instead of Mean-Squared Error (MSE) $\ell_2$ loss as latter has inherent property of generating blurry output as shown in the literature \cite{zhao2015loss}.} and the perceptual loss \cite{johnson2016perceptual} defined as follows.
\begin{align}
    \mathcal{L} = \mathcal{L}_r + \lambda\mathcal{L}_p
\end{align}
Here, $\mathcal{L}_r = \sum\limits_{i} \vert\mathsf{G}_i-\mathsf{S}_i\vert_1$ denotes $\ell_1$ reconstruction loss with $\mathsf{G}_i$ being the ground-truth frame corresponding to the generated frame $\mathsf{S}_i$. $\mathcal{L}_p$ denotes the perceptual loss computed using a pre-trained VGG16 network \cite{johnson2016perceptual}, and $\lambda$ is a hyper-parameter. We use $\lambda$ = 0.2 for all our experiments.

\subsection{Qualitative Results}
Figure~\ref{fig:compare} shows some examples of high frame-rate videos generated using the proposed method and state-of-the-art BIN$_4$~\cite{shen2020blurry} given a low frame-rate video (top row). From Figure~\ref{subfig:compare_a}, it can be seen that our approach is able to tackle the motion blur introduced due to the object motion (car in the bottom left corner for this particular example) along with the blur produced by averaging of consecutive sharp frames. As our approach is extracting information by applying attention on latent representation of input frame, our method is able to deblur and interpolate visually more appealing videos. In Figure~\ref{subfig:compare_b}, the last two frames of middle and bottom row show that the proposed method is able to deblur and interpolate visually good quality frames whereas BIN$_4$ generates a blurry interpolated frame. As the BIN$_4$ utilizes the deblurred frame to interpolate, the error from deblurred frame may propagate during interpolation and hence produce a blurry interpolated frame as shown in Fig~\ref{subfig:compare_b} (middle row, last frame). Our approach overcomes this by generating optimized representation using attention mechanisms, which extracts relevant information from neighbouring frames in the latent space for both deblurring and interpolation.

\begin{figure*}[t]
    \centering
    \includegraphics[width=0.94\linewidth]{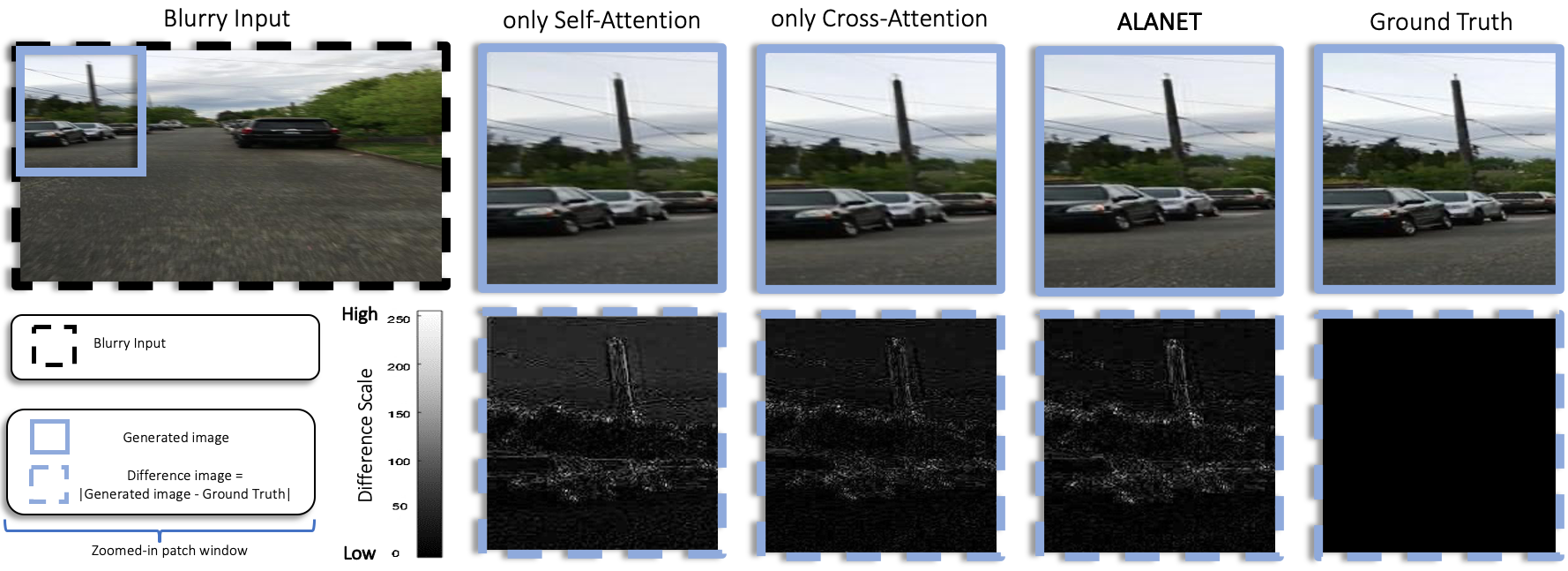}
    \caption{\textit{Ablation study on different attention modules.} Frame generated using different attention mechanisms (top) and the residue image (bottom) computed by taking its difference with the ground-truth frame. Scale for the error range [0. 255] is given on the bottom left. Our proposed ALANET which combines self-attention and cross-attention produces superior results compared to using only one of the attention mechanisms. Results best viewed when zoomed-in.}
    \label{fig:ablation}
\end{figure*}

\subsection{Quantitative Results}
Our proposed method performs joint deblurring and interpolation. There are several methods that only solve the tasks of either deblurring or interpolation. We compare our proposed approach with these state-of-the-art methods that either perform deblurring or interpolation~\cite{jiang2018super, bao2019memc, bao2019depth} given an input blurry video.  We also compare \textbf{ALANET} with two recent approaches where deblurring and interpolation is performed jointly~\cite{jin2018learning, shen2020blurry}. Quantitative result comparison with these baselines are shown in Table~\ref{tab:main}.\bigskip

\noindent \textbf{Results on Adobe240 Dataset.} For deblurring task on Adobe240 dataset, we report a relative improvement of 1.04dB in the average PSNR value and 2.09\% improvement in SSIM metric when compared to ~\cite{shen2020blurry}. Our method achieves 32.98dB average PSNR in interpolation task as opposed 32.51dB reported by state-of-the-art method BIN$_4$~\cite{shen2020blurry}. Overall, for the joint task of deblurring and interpolation the proposed method achieves relative improvement of 2.3\% in average PSNR and 1.04\% in SSIM index against BIN$_4$. It can be observed that BIN$_4$ and \textbf{ALANET} both jointly formulate the deblurring and interpolation tasks which helps to outperform~\cite{jin2018learning}. We again highlight that our method does not know which frames are blurry or where to interpolate, unlike BIN$_4$~\cite{shen2020blurry}.\bigskip

\noindent \textbf{Results on YouTube240 Dataset.} We evaluate the performance of our model trained on Adobe240 dataset for deblurring and interpolation on YouTube240 dataset. For this experiment we crawled 60 videos from YouTube to create this dataset following authors in~\cite{shen2020blurry}. However, we do not have the same set of videos as in~\cite{shen2020blurry} as the list of videos is not publicly available.
From Table~\ref{tab:main}, it can be observed that network trained on Adobe240 performs at-par when evaluated on YouTube240 dataset with average PSNR of  35.89dB and SSIM index of 0.9504 for joint deblurring and interpolation.

\subsection{Ablation Study}
In this section, we investigate the contribution of self-attention and cross-attention in the proposed approach. First, we study the impact of self-attention on video deblurring and interpolation. We remove the cross-attention $\mathcal{M}_C$ terms from ~\eqref{eqn:deblur} and ~\eqref{eqn:interp} and train the network using only self-attention in the latent space. Secondly, we study the impact of cross-attention in absence of self-attention by removing $\mathcal{M}_S$ terms from ~\eqref{eqn:deblur} and ~\eqref{eqn:interp} for training the network. 

Figure~\ref{fig:ablation} presents the qualitative results of the ablation study. It can be observed that the network trained using only self-attention produces inferior results as compared to that of using only cross-attention. The network trained with only self-attention module assumes that all the information to deblur and interpolate resides in a single frame and discards the temporal information available in consecutive frames. This loss in information results in poor quality frame when using only self-attention. On the other hand, using only cross-attention produces better results than using only self-attention module as it exploits the available temporal information by applying cross-attention on latent representation of the consecutive frames. 

The quantitative results of impact of different attention mechanisms are shown in Table~\ref{tab:ablation}. Network trained on only cross-attention achieves improvement of  0.38dB PSNR as compared to using only self-attention for deblurring. However, for interpolation there is improvement of 1.90dB when using only cross-attention as, unlike self-attention, it exploits the temporal information available from neighbouring frames. From Table~\ref{tab:ablation}, we can observe that \textbf{ALANET} performs best as it extracts quality information from the latent representation by exploiting combination of self-attention and cross-attention for deblurring and interpolation.

\begin{table}[t]
    \caption{\textit{Ablation study on attention mechanism.} We evaluate contribution of self-attention and cross-attention for high frame-rate video generation on Adobe240 dataset.}
    \centering
    \begingroup
    \renewcommand{\arraystretch}{1.3}
    \begin{tabular}{c c c c c}
    \toprule[0.2em]
    \multicolumn{1}{c}{\multirow{2}{*}{\textbf{Attention}}} & \multicolumn{2}{l}{\textbf{Deblurring}} & \multicolumn{2}{l}{\textbf{Interpolation}} \\ \cline{2-5} 
    \multicolumn{1}{c}{}                   & PSNR          & SSIM           & PSNR            & SSIM           \\ \hline
    only Self-Attention                    & 31.98         & 0.9373         & 30.87           & 0.9233          \\ \hline
    only Cross-Attention                   & 32.36         & 0.9385         & 32.77           & 0.9340               \\ \hline
    \textbf{ALANET}                        & 33.71         & 0.9429         & 32.98           & 0.9362         \\ \hline
    \label{tab:ablation}
    \end{tabular}
    \endgroup
\end{table}

\section{Conclusion}
We present an Adaptive Latent Attention Network (\textbf{ALANET}) for generating high frame-rate sharp videos with no knowledge that either an input frame is blurry or not. The proposed approach employs self-attention and cross-attention mechanism in the latent representations of input video frames for deblurring and interpolation.
Specifically, the self-attention module extracts information local to the input frame and the cross-attention module exploits the temporal relationship from latent representations of neighbouring frame. Using combination of self-attention and cross-attention our approach is able to generate high frame-rate sharp video.
Experiments on standard datasets show the efficacy of our proposed attention module in task of joint deblurring and interpolation over state-of-the-art methods.

\begin{acks}
This work was partially supported by NSF grants 33397 and 33425, ONR grant N00014-18-1-2252, and a gift from CISCO. We thank Padmaja Jonnalagedda, JVL Venkatesh and JV Megha for valuable discussions and feedback on the paper.

\end{acks}


\bibliographystyle{ACM-Reference-Format}
\balance
\bibliography{sample-base}


\begin{thebibliography}{36}


\ifx \showCODEN    \undefined \def \showCODEN     #1{\unskip}     \fi
\ifx \showDOI      \undefined \def \showDOI       #1{#1}\fi
\ifx \showISBNx    \undefined \def \showISBNx     #1{\unskip}     \fi
\ifx \showISBNxiii \undefined \def \showISBNxiii  #1{\unskip}     \fi
\ifx \showISSN     \undefined \def \showISSN      #1{\unskip}     \fi
\ifx \showLCCN     \undefined \def \showLCCN      #1{\unskip}     \fi
\ifx \shownote     \undefined \def \shownote      #1{#1}          \fi
\ifx \showarticletitle \undefined \def \showarticletitle #1{#1}   \fi
\ifx \showURL      \undefined \def \showURL       {\relax}        \fi
\providecommand\bibfield[2]{#2}
\providecommand\bibinfo[2]{#2}
\providecommand\natexlab[1]{#1}
\providecommand\showeprint[2][]{arXiv:#2}

\bibitem[\protect\citeauthoryear{Aich, Gupta, Panda, Hyder, Asif, and
  Roy-Chowdhury}{Aich et~al\mbox{.}}{2020}]%
        {aich2020non}
\bibfield{author}{\bibinfo{person}{Abhishek Aich}, \bibinfo{person}{Akash
  Gupta}, \bibinfo{person}{Rameswar Panda}, \bibinfo{person}{Rakib Hyder},
  \bibinfo{person}{M~Salman Asif}, {and} \bibinfo{person}{Amit~K
  Roy-Chowdhury}.} \bibinfo{year}{2020}\natexlab{}.
\newblock \showarticletitle{Non-Adversarial Video Synthesis with Learned
  Priors}.
\newblock \bibinfo{journal}{\emph{arXiv preprint arXiv:2003.09565}}
  (\bibinfo{year}{2020}).
\newblock


\bibitem[\protect\citeauthoryear{Ba, Mnih, and Kavukcuoglu}{Ba
  et~al\mbox{.}}{2014}]%
        {ba2014multiple}
\bibfield{author}{\bibinfo{person}{Jimmy Ba}, \bibinfo{person}{Volodymyr Mnih},
  {and} \bibinfo{person}{Koray Kavukcuoglu}.} \bibinfo{year}{2014}\natexlab{}.
\newblock \showarticletitle{Multiple object recognition with visual attention}.
\newblock \bibinfo{journal}{\emph{arXiv preprint arXiv:1412.7755}}
  (\bibinfo{year}{2014}).
\newblock


\bibitem[\protect\citeauthoryear{Bao, Lai, Ma, Zhang, Gao, and Yang}{Bao
  et~al\mbox{.}}{2019a}]%
        {bao2019depth}
\bibfield{author}{\bibinfo{person}{Wenbo Bao}, \bibinfo{person}{Wei-Sheng Lai},
  \bibinfo{person}{Chao Ma}, \bibinfo{person}{Xiaoyun Zhang},
  \bibinfo{person}{Zhiyong Gao}, {and} \bibinfo{person}{Ming-Hsuan Yang}.}
  \bibinfo{year}{2019}\natexlab{a}.
\newblock \showarticletitle{Depth-aware video frame interpolation}. In
  \bibinfo{booktitle}{\emph{Proceedings of the IEEE Conference on Computer
  Vision and Pattern Recognition}}. \bibinfo{pages}{3703--3712}.
\newblock


\bibitem[\protect\citeauthoryear{Bao, Lai, Zhang, Gao, and Yang}{Bao
  et~al\mbox{.}}{2019b}]%
        {bao2019memc}
\bibfield{author}{\bibinfo{person}{Wenbo Bao}, \bibinfo{person}{Wei-Sheng Lai},
  \bibinfo{person}{Xiaoyun Zhang}, \bibinfo{person}{Zhiyong Gao}, {and}
  \bibinfo{person}{Ming-Hsuan Yang}.} \bibinfo{year}{2019}\natexlab{b}.
\newblock \showarticletitle{MEMC-Net: Motion estimation and motion compensation
  driven neural network for video interpolation and enhancement}.
\newblock \bibinfo{journal}{\emph{IEEE transactions on pattern analysis and
  machine intelligence}} (\bibinfo{year}{2019}).
\newblock


\bibitem[\protect\citeauthoryear{Cho, Wang, and Lee}{Cho et~al\mbox{.}}{2012}]%
        {cho2012video}
\bibfield{author}{\bibinfo{person}{Sunghyun Cho}, \bibinfo{person}{Jue Wang},
  {and} \bibinfo{person}{Seungyong Lee}.} \bibinfo{year}{2012}\natexlab{}.
\newblock \showarticletitle{Video deblurring for hand-held cameras using
  patch-based synthesis}.
\newblock \bibinfo{journal}{\emph{ACM Transactions on Graphics (TOG)}}
  \bibinfo{volume}{31}, \bibinfo{number}{4} (\bibinfo{year}{2012}),
  \bibinfo{pages}{1--9}.
\newblock


\bibitem[\protect\citeauthoryear{Choi, Kim, Han, Xu, and Lee}{Choi
  et~al\mbox{.}}{2020}]%
        {choi2020channel}
\bibfield{author}{\bibinfo{person}{Myungsub Choi}, \bibinfo{person}{Heewon
  Kim}, \bibinfo{person}{Bohyung Han}, \bibinfo{person}{Ning Xu}, {and}
  \bibinfo{person}{Kyoung~Mu Lee}.} \bibinfo{year}{2020}\natexlab{}.
\newblock \showarticletitle{Channel attention is all you need for video frame
  interpolation}. AAAI.
\newblock


\bibitem[\protect\citeauthoryear{Hyun~Kim and Mu~Lee}{Hyun~Kim and
  Mu~Lee}{2015}]%
        {hyun2015generalized}
\bibfield{author}{\bibinfo{person}{Tae Hyun~Kim} {and} \bibinfo{person}{Kyoung
  Mu~Lee}.} \bibinfo{year}{2015}\natexlab{}.
\newblock \showarticletitle{Generalized video deblurring for dynamic scenes}.
  In \bibinfo{booktitle}{\emph{Proceedings of the IEEE Conference on Computer
  Vision and Pattern Recognition}}. \bibinfo{pages}{5426--5434}.
\newblock


\bibitem[\protect\citeauthoryear{Hyun~Kim, Mu~Lee, Scholkopf, and
  Hirsch}{Hyun~Kim et~al\mbox{.}}{2017}]%
        {hyun2017online}
\bibfield{author}{\bibinfo{person}{Tae Hyun~Kim}, \bibinfo{person}{Kyoung
  Mu~Lee}, \bibinfo{person}{Bernhard Scholkopf}, {and} \bibinfo{person}{Michael
  Hirsch}.} \bibinfo{year}{2017}\natexlab{}.
\newblock \showarticletitle{Online video deblurring via dynamic temporal
  blending network}. In \bibinfo{booktitle}{\emph{Proceedings of the IEEE
  International Conference on Computer Vision}}. \bibinfo{pages}{4038--4047}.
\newblock


\bibitem[\protect\citeauthoryear{Jiang, Sun, Jampani, Yang, Learned-Miller, and
  Kautz}{Jiang et~al\mbox{.}}{2018}]%
        {jiang2018super}
\bibfield{author}{\bibinfo{person}{Huaizu Jiang}, \bibinfo{person}{Deqing Sun},
  \bibinfo{person}{Varun Jampani}, \bibinfo{person}{Ming-Hsuan Yang},
  \bibinfo{person}{Erik Learned-Miller}, {and} \bibinfo{person}{Jan Kautz}.}
  \bibinfo{year}{2018}\natexlab{}.
\newblock \showarticletitle{Super slomo: High quality estimation of multiple
  intermediate frames for video interpolation}. In
  \bibinfo{booktitle}{\emph{Proceedings of the IEEE Conference on Computer
  Vision and Pattern Recognition}}. \bibinfo{pages}{9000--9008}.
\newblock


\bibitem[\protect\citeauthoryear{Jin, Meishvili, and Favaro}{Jin
  et~al\mbox{.}}{2018}]%
        {jin2018learning}
\bibfield{author}{\bibinfo{person}{Meiguang Jin}, \bibinfo{person}{Givi
  Meishvili}, {and} \bibinfo{person}{Paolo Favaro}.}
  \bibinfo{year}{2018}\natexlab{}.
\newblock \showarticletitle{Learning to extract a video sequence from a single
  motion-blurred image}. In \bibinfo{booktitle}{\emph{Proceedings of the IEEE
  Conference on Computer Vision and Pattern Recognition}}.
  \bibinfo{pages}{6334--6342}.
\newblock


\bibitem[\protect\citeauthoryear{Johnson, Alahi, and Fei-Fei}{Johnson
  et~al\mbox{.}}{2016}]%
        {johnson2016perceptual}
\bibfield{author}{\bibinfo{person}{Justin Johnson}, \bibinfo{person}{Alexandre
  Alahi}, {and} \bibinfo{person}{Li Fei-Fei}.} \bibinfo{year}{2016}\natexlab{}.
\newblock \showarticletitle{Perceptual losses for real-time style transfer and
  super-resolution}. In \bibinfo{booktitle}{\emph{European conference on
  computer vision}}. Springer, \bibinfo{pages}{694--711}.
\newblock


\bibitem[\protect\citeauthoryear{Kim, Nah, and Lee}{Kim et~al\mbox{.}}{2016}]%
        {kim2016dynamic}
\bibfield{author}{\bibinfo{person}{Tae~Hyun Kim}, \bibinfo{person}{Seungjun
  Nah}, {and} \bibinfo{person}{Kyoung~Mu Lee}.}
  \bibinfo{year}{2016}\natexlab{}.
\newblock \showarticletitle{Dynamic scene deblurring using a locally adaptive
  linear blur model}.
\newblock \bibinfo{journal}{\emph{arXiv preprint arXiv:1603.04265}}
  (\bibinfo{year}{2016}).
\newblock


\bibitem[\protect\citeauthoryear{Kim, Nah, and Lee}{Kim et~al\mbox{.}}{2017}]%
        {kim2017dynamic}
\bibfield{author}{\bibinfo{person}{Tae~Hyun Kim}, \bibinfo{person}{Seungjun
  Nah}, {and} \bibinfo{person}{Kyoung~Mu Lee}.}
  \bibinfo{year}{2017}\natexlab{}.
\newblock \showarticletitle{Dynamic video deblurring using a locally adaptive
  blur model}.
\newblock \bibinfo{journal}{\emph{IEEE transactions on pattern analysis and
  machine intelligence}} \bibinfo{volume}{40}, \bibinfo{number}{10}
  (\bibinfo{year}{2017}), \bibinfo{pages}{2374--2387}.
\newblock


\bibitem[\protect\citeauthoryear{Kingma and Ba}{Kingma and Ba}{2014}]%
        {kingma2014adam}
\bibfield{author}{\bibinfo{person}{Diederik~P Kingma} {and}
  \bibinfo{person}{Jimmy Ba}.} \bibinfo{year}{2014}\natexlab{}.
\newblock \showarticletitle{Adam: A method for stochastic optimization}.
\newblock \bibinfo{journal}{\emph{arXiv preprint arXiv:1412.6980}}
  (\bibinfo{year}{2014}).
\newblock


\bibitem[\protect\citeauthoryear{Liu, Yeh, Tang, Liu, and Agarwala}{Liu
  et~al\mbox{.}}{2017}]%
        {liu2017video}
\bibfield{author}{\bibinfo{person}{Ziwei Liu}, \bibinfo{person}{Raymond~A Yeh},
  \bibinfo{person}{Xiaoou Tang}, \bibinfo{person}{Yiming Liu}, {and}
  \bibinfo{person}{Aseem Agarwala}.} \bibinfo{year}{2017}\natexlab{}.
\newblock \showarticletitle{Video frame synthesis using deep voxel flow}. In
  \bibinfo{booktitle}{\emph{Proceedings of the IEEE International Conference on
  Computer Vision}}. \bibinfo{pages}{4463--4471}.
\newblock


\bibitem[\protect\citeauthoryear{Long, Kneip, Alvarez, Li, Zhang, and Yu}{Long
  et~al\mbox{.}}{2016}]%
        {long2016learning}
\bibfield{author}{\bibinfo{person}{Gucan Long}, \bibinfo{person}{Laurent
  Kneip}, \bibinfo{person}{Jose~M Alvarez}, \bibinfo{person}{Hongdong Li},
  \bibinfo{person}{Xiaohu Zhang}, {and} \bibinfo{person}{Qifeng Yu}.}
  \bibinfo{year}{2016}\natexlab{}.
\newblock \showarticletitle{Learning image matching by simply watching video}.
  In \bibinfo{booktitle}{\emph{European Conference on Computer Vision}}.
  Springer, \bibinfo{pages}{434--450}.
\newblock


\bibitem[\protect\citeauthoryear{Mahajan, Huang, Matusik, Ramamoorthi, and
  Belhumeur}{Mahajan et~al\mbox{.}}{2009}]%
        {mahajan2009moving}
\bibfield{author}{\bibinfo{person}{Dhruv Mahajan}, \bibinfo{person}{Fu-Chung
  Huang}, \bibinfo{person}{Wojciech Matusik}, \bibinfo{person}{Ravi
  Ramamoorthi}, {and} \bibinfo{person}{Peter Belhumeur}.}
  \bibinfo{year}{2009}\natexlab{}.
\newblock \showarticletitle{Moving gradients: a path-based method for plausible
  image interpolation}.
\newblock \bibinfo{journal}{\emph{ACM Transactions on Graphics (TOG)}}
  \bibinfo{volume}{28}, \bibinfo{number}{3} (\bibinfo{year}{2009}),
  \bibinfo{pages}{1--11}.
\newblock


\bibitem[\protect\citeauthoryear{Nah, Hyun~Kim, and Mu~Lee}{Nah
  et~al\mbox{.}}{2017}]%
        {nah2017deep}
\bibfield{author}{\bibinfo{person}{Seungjun Nah}, \bibinfo{person}{Tae
  Hyun~Kim}, {and} \bibinfo{person}{Kyoung Mu~Lee}.}
  \bibinfo{year}{2017}\natexlab{}.
\newblock \showarticletitle{Deep multi-scale convolutional neural network for
  dynamic scene deblurring}. In \bibinfo{booktitle}{\emph{Proceedings of the
  IEEE Conference on Computer Vision and Pattern Recognition}}.
  \bibinfo{pages}{3883--3891}.
\newblock


\bibitem[\protect\citeauthoryear{Niklaus, Mai, and Liu}{Niklaus
  et~al\mbox{.}}{2017a}]%
        {niklaus2017video}
\bibfield{author}{\bibinfo{person}{Simon Niklaus}, \bibinfo{person}{Long Mai},
  {and} \bibinfo{person}{Feng Liu}.} \bibinfo{year}{2017}\natexlab{a}.
\newblock \showarticletitle{Video frame interpolation via adaptive
  convolution}. In \bibinfo{booktitle}{\emph{Proceedings of the IEEE Conference
  on Computer Vision and Pattern Recognition}}. \bibinfo{pages}{670--679}.
\newblock


\bibitem[\protect\citeauthoryear{Niklaus, Mai, and Liu}{Niklaus
  et~al\mbox{.}}{2017b}]%
        {niklaus2017video2}
\bibfield{author}{\bibinfo{person}{Simon Niklaus}, \bibinfo{person}{Long Mai},
  {and} \bibinfo{person}{Feng Liu}.} \bibinfo{year}{2017}\natexlab{b}.
\newblock \showarticletitle{Video frame interpolation via adaptive separable
  convolution}. In \bibinfo{booktitle}{\emph{Proceedings of the IEEE
  International Conference on Computer Vision}}. \bibinfo{pages}{261--270}.
\newblock


\bibitem[\protect\citeauthoryear{Nimisha, Kumar~Singh, and Rajagopalan}{Nimisha
  et~al\mbox{.}}{2017}]%
        {nimisha2017blur}
\bibfield{author}{\bibinfo{person}{Thekke~Madam Nimisha},
  \bibinfo{person}{Akash Kumar~Singh}, {and} \bibinfo{person}{Ambasamudram~N
  Rajagopalan}.} \bibinfo{year}{2017}\natexlab{}.
\newblock \showarticletitle{Blur-invariant deep learning for blind-deblurring}.
  In \bibinfo{booktitle}{\emph{Proceedings of the IEEE International Conference
  on Computer Vision}}. \bibinfo{pages}{4752--4760}.
\newblock


\bibitem[\protect\citeauthoryear{Paszke, Gross, Chintala, Chanan, Yang, DeVito,
  Lin, Desmaison, Antiga, and Lerer}{Paszke et~al\mbox{.}}{2017}]%
        {paszke2017automatic}
\bibfield{author}{\bibinfo{person}{Adam Paszke}, \bibinfo{person}{Sam Gross},
  \bibinfo{person}{Soumith Chintala}, \bibinfo{person}{Gregory Chanan},
  \bibinfo{person}{Edward Yang}, \bibinfo{person}{Zachary DeVito},
  \bibinfo{person}{Zeming Lin}, \bibinfo{person}{Alban Desmaison},
  \bibinfo{person}{Luca Antiga}, {and} \bibinfo{person}{Adam Lerer}.}
  \bibinfo{year}{2017}\natexlab{}.
\newblock \showarticletitle{Automatic {D}ifferentiation in {PyTorch}}. In
  \bibinfo{booktitle}{\emph{NIPS AutoDiff Workshop}}.
\newblock


\bibitem[\protect\citeauthoryear{Raskar, Agrawal, and Tumblin}{Raskar
  et~al\mbox{.}}{2006}]%
        {raskar2006coded}
\bibfield{author}{\bibinfo{person}{Ramesh Raskar}, \bibinfo{person}{Amit
  Agrawal}, {and} \bibinfo{person}{Jack Tumblin}.}
  \bibinfo{year}{2006}\natexlab{}.
\newblock \showarticletitle{Coded exposure photography: motion deblurring using
  fluttered shutter}.
\newblock In \bibinfo{booktitle}{\emph{ACM SIGGRAPH 2006 Papers}}.
  \bibinfo{pages}{795--804}.
\newblock


\bibitem[\protect\citeauthoryear{Ronneberger, Fischer, and Brox}{Ronneberger
  et~al\mbox{.}}{2015}]%
        {ronneberger2015u}
\bibfield{author}{\bibinfo{person}{Olaf Ronneberger}, \bibinfo{person}{Philipp
  Fischer}, {and} \bibinfo{person}{Thomas Brox}.}
  \bibinfo{year}{2015}\natexlab{}.
\newblock \showarticletitle{U-net: Convolutional networks for biomedical image
  segmentation}. In \bibinfo{booktitle}{\emph{International Conference on
  Medical image computing and computer-assisted intervention}}. Springer,
  \bibinfo{pages}{234--241}.
\newblock


\bibitem[\protect\citeauthoryear{Shen, Bao, Zhai, Chen, Min, and Gao}{Shen
  et~al\mbox{.}}{2020}]%
        {shen2020blurry}
\bibfield{author}{\bibinfo{person}{Wang Shen}, \bibinfo{person}{Wenbo Bao},
  \bibinfo{person}{Guangtao Zhai}, \bibinfo{person}{Li Chen},
  \bibinfo{person}{Xiongkuo Min}, {and} \bibinfo{person}{Zhiyong Gao}.}
  \bibinfo{year}{2020}\natexlab{}.
\newblock \bibinfo{title}{Blurry Video Frame Interpolation}.
\newblock
\newblock
\showeprint[arxiv]{cs.CV/2002.12259}


\bibitem[\protect\citeauthoryear{Su, Delbracio, Wang, Sapiro, Heidrich, and
  Wang}{Su et~al\mbox{.}}{2017}]%
        {su2017deep}
\bibfield{author}{\bibinfo{person}{Shuochen Su}, \bibinfo{person}{Mauricio
  Delbracio}, \bibinfo{person}{Jue Wang}, \bibinfo{person}{Guillermo Sapiro},
  \bibinfo{person}{Wolfgang Heidrich}, {and} \bibinfo{person}{Oliver Wang}.}
  \bibinfo{year}{2017}\natexlab{}.
\newblock \showarticletitle{Deep video deblurring for hand-held cameras}. In
  \bibinfo{booktitle}{\emph{Proceedings of the IEEE Conference on Computer
  Vision and Pattern Recognition}}. \bibinfo{pages}{1279--1288}.
\newblock


\bibitem[\protect\citeauthoryear{Tao, Gao, Shen, Wang, and Jia}{Tao
  et~al\mbox{.}}{2018}]%
        {tao2018scale}
\bibfield{author}{\bibinfo{person}{Xin Tao}, \bibinfo{person}{Hongyun Gao},
  \bibinfo{person}{Xiaoyong Shen}, \bibinfo{person}{Jue Wang}, {and}
  \bibinfo{person}{Jiaya Jia}.} \bibinfo{year}{2018}\natexlab{}.
\newblock \showarticletitle{Scale-recurrent network for deep image deblurring}.
  In \bibinfo{booktitle}{\emph{Proceedings of the IEEE Conference on Computer
  Vision and Pattern Recognition}}. \bibinfo{pages}{8174--8182}.
\newblock


\bibitem[\protect\citeauthoryear{Telleen, Sullivan, Yee, Wang, Gunawardane,
  Collins, and Davis}{Telleen et~al\mbox{.}}{2007}]%
        {telleen2007synthetic}
\bibfield{author}{\bibinfo{person}{Jacob Telleen}, \bibinfo{person}{Anne
  Sullivan}, \bibinfo{person}{Jerry Yee}, \bibinfo{person}{Oliver Wang},
  \bibinfo{person}{Prabath Gunawardane}, \bibinfo{person}{Ian Collins}, {and}
  \bibinfo{person}{James Davis}.} \bibinfo{year}{2007}\natexlab{}.
\newblock \showarticletitle{Synthetic shutter speed imaging}. In
  \bibinfo{booktitle}{\emph{Computer Graphics Forum}},
  Vol.~\bibinfo{volume}{26}. Wiley Online Library, \bibinfo{pages}{591--598}.
\newblock


\bibitem[\protect\citeauthoryear{Vaswani, Shazeer, Parmar, Uszkoreit, Jones,
  Gomez, Kaiser, and Polosukhin}{Vaswani et~al\mbox{.}}{2017}]%
        {vaswani2017attention}
\bibfield{author}{\bibinfo{person}{Ashish Vaswani}, \bibinfo{person}{Noam
  Shazeer}, \bibinfo{person}{Niki Parmar}, \bibinfo{person}{Jakob Uszkoreit},
  \bibinfo{person}{Llion Jones}, \bibinfo{person}{Aidan~N Gomez},
  \bibinfo{person}{{\L}ukasz Kaiser}, {and} \bibinfo{person}{Illia
  Polosukhin}.} \bibinfo{year}{2017}\natexlab{}.
\newblock \showarticletitle{Attention is all you need}. In
  \bibinfo{booktitle}{\emph{Advances in neural information processing
  systems}}. \bibinfo{pages}{5998--6008}.
\newblock


\bibitem[\protect\citeauthoryear{Wang, Chan, Yu, Dong, and Change~Loy}{Wang
  et~al\mbox{.}}{2019}]%
        {wang2019edvr}
\bibfield{author}{\bibinfo{person}{Xintao Wang}, \bibinfo{person}{Kelvin~CK
  Chan}, \bibinfo{person}{Ke Yu}, \bibinfo{person}{Chao Dong}, {and}
  \bibinfo{person}{Chen Change~Loy}.} \bibinfo{year}{2019}\natexlab{}.
\newblock \showarticletitle{Edvr: Video restoration with enhanced deformable
  convolutional networks}. In \bibinfo{booktitle}{\emph{Proceedings of the IEEE
  Conference on Computer Vision and Pattern Recognition Workshops}}.
  \bibinfo{pages}{0--0}.
\newblock


\bibitem[\protect\citeauthoryear{Wu, Yu, Liu, Chandraker, and Wang}{Wu
  et~al\mbox{.}}{2020}]%
        {wu2020david}
\bibfield{author}{\bibinfo{person}{Junru Wu}, \bibinfo{person}{Xiang Yu},
  \bibinfo{person}{Ding Liu}, \bibinfo{person}{Manmohan Chandraker}, {and}
  \bibinfo{person}{Zhangyang Wang}.} \bibinfo{year}{2020}\natexlab{}.
\newblock \showarticletitle{DAVID: Dual-Attentional Video Deblurring}. In
  \bibinfo{booktitle}{\emph{The IEEE Winter Conference on Applications of
  Computer Vision}}. \bibinfo{pages}{2376--2385}.
\newblock


\bibitem[\protect\citeauthoryear{Zhang, Goodfellow, Metaxas, and Odena}{Zhang
  et~al\mbox{.}}{2018a}]%
        {zhang2018self}
\bibfield{author}{\bibinfo{person}{Han Zhang}, \bibinfo{person}{Ian
  Goodfellow}, \bibinfo{person}{Dimitris Metaxas}, {and}
  \bibinfo{person}{Augustus Odena}.} \bibinfo{year}{2018}\natexlab{a}.
\newblock \showarticletitle{Self-attention generative adversarial networks}.
\newblock \bibinfo{journal}{\emph{arXiv preprint arXiv:1805.08318}}
  (\bibinfo{year}{2018}).
\newblock


\bibitem[\protect\citeauthoryear{Zhang, Li, Li, Wang, Zhong, and Fu}{Zhang
  et~al\mbox{.}}{2018b}]%
        {zhang2018image}
\bibfield{author}{\bibinfo{person}{Yulun Zhang}, \bibinfo{person}{Kunpeng Li},
  \bibinfo{person}{Kai Li}, \bibinfo{person}{Lichen Wang},
  \bibinfo{person}{Bineng Zhong}, {and} \bibinfo{person}{Yun Fu}.}
  \bibinfo{year}{2018}\natexlab{b}.
\newblock \showarticletitle{Image super-resolution using very deep residual
  channel attention networks}. In \bibinfo{booktitle}{\emph{Proceedings of the
  European Conference on Computer Vision (ECCV)}}. \bibinfo{pages}{286--301}.
\newblock


\bibitem[\protect\citeauthoryear{Zhao, Gallo, Frosio, and Kautz}{Zhao
  et~al\mbox{.}}{2015}]%
        {zhao2015loss}
\bibfield{author}{\bibinfo{person}{Hang Zhao}, \bibinfo{person}{Orazio Gallo},
  \bibinfo{person}{Iuri Frosio}, {and} \bibinfo{person}{Jan Kautz}.}
  \bibinfo{year}{2015}\natexlab{}.
\newblock \showarticletitle{Loss functions for neural networks for image
  processing}.
\newblock \bibinfo{journal}{\emph{arXiv preprint arXiv:1511.08861}}
  (\bibinfo{year}{2015}).
\newblock


\bibitem[\protect\citeauthoryear{Zhou, Zhang, Pan, Xie, Zuo, and Ren}{Zhou
  et~al\mbox{.}}{2019}]%
        {zhou2019spatio}
\bibfield{author}{\bibinfo{person}{Shangchen Zhou}, \bibinfo{person}{Jiawei
  Zhang}, \bibinfo{person}{Jinshan Pan}, \bibinfo{person}{Haozhe Xie},
  \bibinfo{person}{Wangmeng Zuo}, {and} \bibinfo{person}{Jimmy Ren}.}
  \bibinfo{year}{2019}\natexlab{}.
\newblock \showarticletitle{Spatio-temporal filter adaptive network for video
  deblurring}. In \bibinfo{booktitle}{\emph{Proceedings of the IEEE
  International Conference on Computer Vision}}. \bibinfo{pages}{2482--2491}.
\newblock


\bibitem[\protect\citeauthoryear{Zitnick, Kang, Uyttendaele, Winder, and
  Szeliski}{Zitnick et~al\mbox{.}}{2004}]%
        {zitnick2004high}
\bibfield{author}{\bibinfo{person}{C~Lawrence Zitnick},
  \bibinfo{person}{Sing~Bing Kang}, \bibinfo{person}{Matthew Uyttendaele},
  \bibinfo{person}{Simon Winder}, {and} \bibinfo{person}{Richard Szeliski}.}
  \bibinfo{year}{2004}\natexlab{}.
\newblock \showarticletitle{High-quality video view interpolation using a
  layered representation}.
\newblock \bibinfo{journal}{\emph{ACM transactions on graphics (TOG)}}
  \bibinfo{volume}{23}, \bibinfo{number}{3} (\bibinfo{year}{2004}),
  \bibinfo{pages}{600--608}.
\newblock


\end{thebibliography}

\appendix

\end{document}